\documentclass[fleqn,10pt]{wlscirep}
\usepackage[utf8]{inputenc}
\usepackage[T1]{fontenc}
\usepackage{multirow}
\usepackage{subcaption}

\usepackage{float} 
\usepackage{rotating}  
\title{SuperiorGAT: Graph Attention Networks for Sparse LiDAR Point Cloud Reconstruction in Autonomous Systems}

\author[1,*]{Khalfalla Awedat}
\author[2]{Mohamed Abidalrekab}
\author[3]{Gurcan Comert}
\author[4]{Mustafa Ayad}
\affil{Computer Information Technology Department, SUNY Morrisville College, Morrisville, NY, USA}
\affil[2]{Electrical and Computer Engineering, Portland State University, Portland, OR, USA}
\affil[3]{Computational Data Science and Engineering Department, North Carolina A\&T State University, Greensboro, NC, USA}
\affil[4]{Electrical and Computer Engineering Department, SUNY Oswego,Oswego, NY, USA}

\affil[*]{awedatk@morrisville.edu}

\begin{abstract}
LiDAR-based perception in autonomous systems is fundamentally constrained by fixed vertical beam resolution and is further degraded by structured beam dropout caused by occlusions or reduced-cost sensing hardware. This paper introduces \textbf{SuperiorGAT}, a graph attention--based framework for reconstructing missing elevation information in sparse LiDAR point clouds under structured beam loss. By modeling LiDAR scans as beam-aware graphs and augmenting standard graph attention networks with gated residual fusion and lightweight feed-forward refinement, the proposed approach improves vertical reconstruction accuracy without increasing network depth.

The effectiveness of SuperiorGAT is evaluated through extensive experiments on multiple KITTI environments, including \textbf{Person}, \textbf{Road}, \textbf{Campus}, and \textbf{City}, as well as cross-dataset validation on nuScenes with lower vertical resolution. Robustness is assessed under increasing structured beam dropout, demonstrating that SuperiorGAT consistently achieves lower reconstruction error and improved geometric consistency compared to interpolation-based methods, PointNet-based models, and deeper GAT baselines. Qualitative X--Z projection analyses further confirm the model’s ability to preserve structural continuity with minimal vertical distortion. Overall, the results indicate that targeted architectural refinement provides a computationally efficient solution for enhancing LiDAR vertical reconstruction without reliance on additional sensor modalities or hardware upgrades.
\end{abstract}

\begin{document}

\flushbottom
\maketitle
\thispagestyle{empty}

\section*{Introduction}

Autonomous vehicles rely critically on Light Detection and Ranging (LiDAR) sensors for 3D environmental perception, enabling precise object detection, localization, and mapping essential for safe navigation \cite{zhang2022,geiger2013}. Despite their fundamental importance, LiDAR systems face a critical vulnerability: performance severely degrades under structured data loss, particularly beam dropout caused by hardware malfunctions or environmental occlusions, which disrupts spatial continuity and compromises downstream tasks such as object detection and path planning \cite{zhang2022,behley2019}. This challenge is particularly acute for cost-effective LiDAR systems that employ fewer scanning beams (typically 16-32 beams compared to 64+ in premium systems), inherently producing sparse point clouds that demand robust reconstruction methods to achieve the high-resolution performance required for autonomous operation \cite{dosovitskiy2017}.

Current reconstruction approaches face a fundamental trade-off between accuracy and computational efficiency, limiting their practical deployment. Compressive sensing (CS) methods provide theoretical guarantees for sparse signal recovery but depend on computationally intensive iterative solvers that render them impractical for the real-time constraints of autonomous systems \cite{candes2006,donoho2006}. Convolutional neural networks (CNNs), particularly voxel-based architectures, demonstrate effectiveness in processing point clouds but necessitate dense 3D projections, resulting in inefficient memory usage and computational overhead when applied to inherently sparse LiDAR data \cite{wang2018,hatun2023}.

Recent advances in deep learning, including hierarchical point cloud models that operate directly on unordered point sets, have shown promise but remain inadequate for addressing the structured sparsity patterns characteristic of beam dropout scenarios \cite{qi2017}. While Graph Neural Networks (GNNs), especially Graph Attention Networks (GATs), have demonstrated significant potential in modeling complex geometric relationships within irregular data structures through sophisticated graph representations \cite{scarselli2009,velickovic2018,turati2022}, standard graph-based approaches, including graph convolutional networks, often fail to effectively prioritize the critical local geometric features essential for high-fidelity reconstruction of sparse LiDAR data \cite{kipf2017}. Our recent conference work demonstrated that multi-layer Graph Attention Networks can effectively reconstruct missing LiDAR elevation values using raw point-cloud geometry alone, without relying on RGB images or temporal information \cite{awedat2025}. While increasing network depth improved reconstruction fidelity, deeper attention stacks introduced higher computational cost and stability limitations. In this work, we move beyond depth-based improvements and focus on architectural refinement to achieve superior performance with a lightweight design.

To address these fundamental limitations, we propose SuperiorGAT, a novel graph attention-based framework specifically engineered to reconstruct sparse LiDAR point clouds with exceptional fidelity while maintaining the computational efficiency necessary for real-time deployment in autonomous systems \cite{hatun2023,turati2022}. Our approach directly tackles the structured sparsity challenge through several key innovations. SuperiorGAT implements a realistic beam dropout simulation framework that accurately mimics hardware faults observed in real-world LiDAR deployments, providing a robust testing environment for autonomous vehicle applications \cite{zhang2022,behley2019}. The system constructs optimized k-nearest neighbor graphs to effectively represent sparse point cloud structures while incorporating specialized beam index features that enhance both spatial and structural contextual understanding \cite{turati2022}.

The model architecture features a carefully designed lightweight framework incorporating multiple graph attention layers with multi-head attention mechanisms and residual connections, enabling highly efficient processing of sparse data without the computational burden of iterative CS optimization or the memory overhead of dense CNN projections \cite{velickovic2018}. By directly addressing structured sparsity patterns, SuperiorGAT provides a scalable and practical solution for autonomous driving applications, successfully bridging the critical gap between high-resolution reconstruction accuracy and real-time computational efficiency. Our primary contributions to the field include:
\begin{enumerate}
\item A comprehensive and realistic simulation framework for structured beam dropout that accurately reflects hardware faults and environmental conditions encountered in real-world LiDAR systems.
\item A novel graph attention-based reconstruction architecture that dynamically optimizes point-to-point interactions through learned attention weights, effectively preserving fine-grained structural details essential for autonomous navigation.
\item A computationally efficient framework specifically designed for real-time deployment constraints, offering significant potential for practical integration in autonomous vehicle systems.
\end{enumerate}

\begin{figure}[h]
\centering
\includegraphics[width=0.75\textwidth]{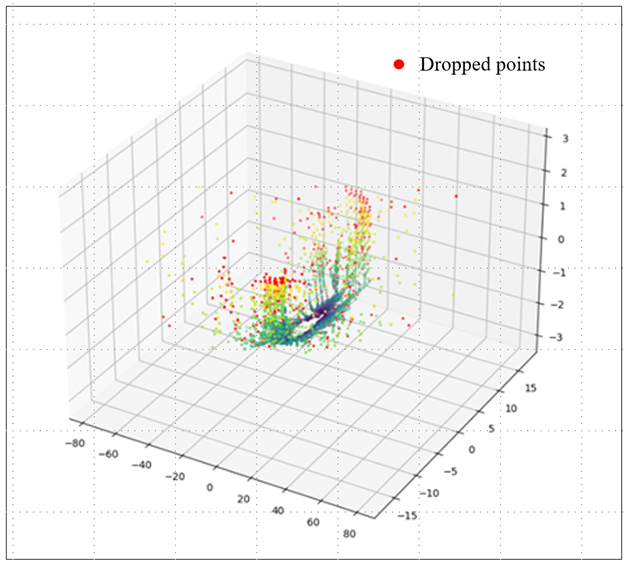}
\caption{LiDAR point cloud with beam dropout (KITTI Residential dataset). Original 64-beam point cloud (colored by z-coordinate) with 25\% beam dropout points in red, simulating missing z-coordinates due to hardware or environmental faults. SuperiorGAT reconstructs these z-values for autonomous driving.}
\label{fig:dropout}
\end{figure}

The remainder of this paper is structured as follows. Section 2 provides a comprehensive review of related work in LiDAR point cloud reconstruction and graph-based methodologies. Section 3 presents detailed technical specifications of the SuperiorGAT methodology, including data preprocessing pipelines and architectural design principles. Section 4 describes our experimental framework and evaluation protocols, followed by a comprehensive analysis of the results and performance in Section 5. Section 6 discusses the broader implications of our findings, practical limitations, and potential applications, while Section 7 concludes with future research directions and development paths.

\section*{Related Work}

Light Detection and Ranging (LiDAR) technology is a cornerstone of 3D perception for autonomous vehicles, enabling tasks like object detection and mapping through precise geometric data \cite{geiger2013,behley2019}. Yet, sparse point clouds—caused by low-resolution sensors or beam dropout from environmental factors such as rain or hardware faults—complicate accurate reconstruction \cite{zhang2021}. As illustrated in Figure \ref{fig:dropout}, beam dropout predominantly disrupts elevation (z-coordinates), while x and y coordinates, derived from azimuthal scans, remain stable, underscoring the need for targeted z-reconstruction methods \cite{eskandar2023,glennie2013}. This vertical sparsity stems from the limited number of laser channels in affordable LiDAR systems, a challenge evident in datasets like KITTI \cite{geiger2013}.

Conventional reconstruction techniques include interpolation and compressive sensing (CS). Interpolation methods, such as linear or nearest-neighbor approaches, estimate missing points based on proximity but often smooth out critical geometric features, particularly in complex scenes \cite{candes2006,donoho2006,rusu2009}. Advanced techniques like spline interpolation reduce this issue but can introduce artifacts in regions with sharp elevation shifts \cite{rusu2009}. CS offers a theoretical framework for recovering sparse signals but demands computationally intensive iterative processes, rendering it unsuitable for real-time applications where sub-100ms inference is critical \cite{candes2006,donoho2006,velickovic2018}. These limitations highlight the demand for efficient solutions tailored to LiDAR’s structured sparsity.

Deep learning has transformed point cloud processing, with Convolutional Neural Networks (CNNs) like VoxelNet converting sparse data into dense voxel grids for tasks such as 3D object detection \cite{wang2018}. However, this densification escalates computational costs and erodes fine details in sparse LiDAR data \cite{wang2018,you2023}. VoxelNet’s cubic scaling with resolution contrasts with PointNet++, which uses hierarchical sampling to manage irregular data, yet both falter with structured dropout patterns due to static feature aggregation \cite{qi2017,graham2018}. PointNet and its extensions leverage permutation invariance but struggle to adapt to the specific sparsity of beam dropout \cite{qi2017,eskandar2023}.

Recent advances in LiDAR super-resolution target vertical resolution enhancement, focusing on z-coordinates for improved pedestrian detection \cite{you2023,eskandar2023}. You and Kim (2023) introduced a 2D range image-based upsampling technique, interpolating x, y, z, and intensity to boost 3D detection accuracy, while Eskandar et al. (2023) proposed HALS, a height-aware model using polar coordinate regression to excel on KITTI data \cite{you2023,eskandar2023}. These approaches affirm z-reconstruction’s importance, given the reliability of x and y coordinates in LiDAR scans \cite{zhang2021}, though their reliance on dense architectures limits real-time viability.

Graph Neural Networks (GNNs) provide a versatile framework for irregular data, modeling points as nodes and relationships as edges \cite{scarselli2009}. Graph Convolutional Networks (GCNs) apply fixed-weight convolutions, effective for node classification but less adept at prioritizing sparse LiDAR geometries \cite{kipf2017}. Enhanced GNNs incorporate edge attributes like distances, enhancing spatial modeling \cite{wang2019,turati2022}. Graph Attention Networks (GATs), pioneered by Veličković et al, introduce dynamic weighting via attention mechanisms, drawing from neural machine translation insights \cite{velickovic2018,bahdanau2015}. Though successful in network analysis, GATs remain underexplored for LiDAR, where their adaptive aggregation suits sparse geometries \cite{wang2019}.
Recent work in graph neural networks has explored architectural elements that enhance stability and expressive capacity beyond the standard GAT formulation. In particular, gated skip connections have been shown to improve gradient flow and help models balance raw node features with attention-refined representations \cite{li2019deepgcns,bresson2017residual}. In our prior work \cite{awedat2025}, we applied a multi-layer GAT architecture to LiDAR beam reconstruction and showed that attention-based message passing is effective for recovering missing elevation data. The present work extends this direction by introducing gated residual fusion and feed-forward refinement, enabling improved reconstruction without increasing graph depth. Likewise, several graph transformer variants incorporate feed-forward sublayers similar to those used in sequence transformers to increase nonlinear capacity and improve local geometric modeling \cite{dwivedi2020generalization}. These ideas motivate our adoption of a lightweight gated residual pathway and a transformer-style feed-forward block, adapted specifically to the characteristics of sparse LiDAR reconstruction.

SuperiorGAT advances GATs by optimizing for LiDAR z-reconstruction, integrating beam index features to reflect scanning patterns and employing a lightweight design with multi-head attention and residuals \cite{velickovic2018,wang2019}. Unlike previous methods, it tackles structured beam dropout (Figure \ref{fig:dropout}) with a focus on z-coordinates, offering a balance of accuracy and efficiency for autonomous driving. This sets it apart from dense CNN approaches \cite{wang2018}, PointNet variants \cite{qi2017}, and super-resolution techniques \cite{you2023,eskandar2023}, establishing a novel real-time solution.

\section*{Graph-Based Reconstruction Methodology}

\subsection*{Graph Neural Networks (GNNs)}

Graph Neural Networks (GNNs) represent a powerful class of deep learning models designed to perform inference on data described by graphs. By leveraging the inherent relational structure of graph data, GNNs overcome a key limitation of traditional neural networks, which require input data to exist in an independent and identically distributed (i.i.d.) or grid-like structure (e.g., images) \cite{scarselli2009, wu2021survey}.

\begin{figure}[ht]
    \centering
    \includegraphics[width=0.75\textwidth]{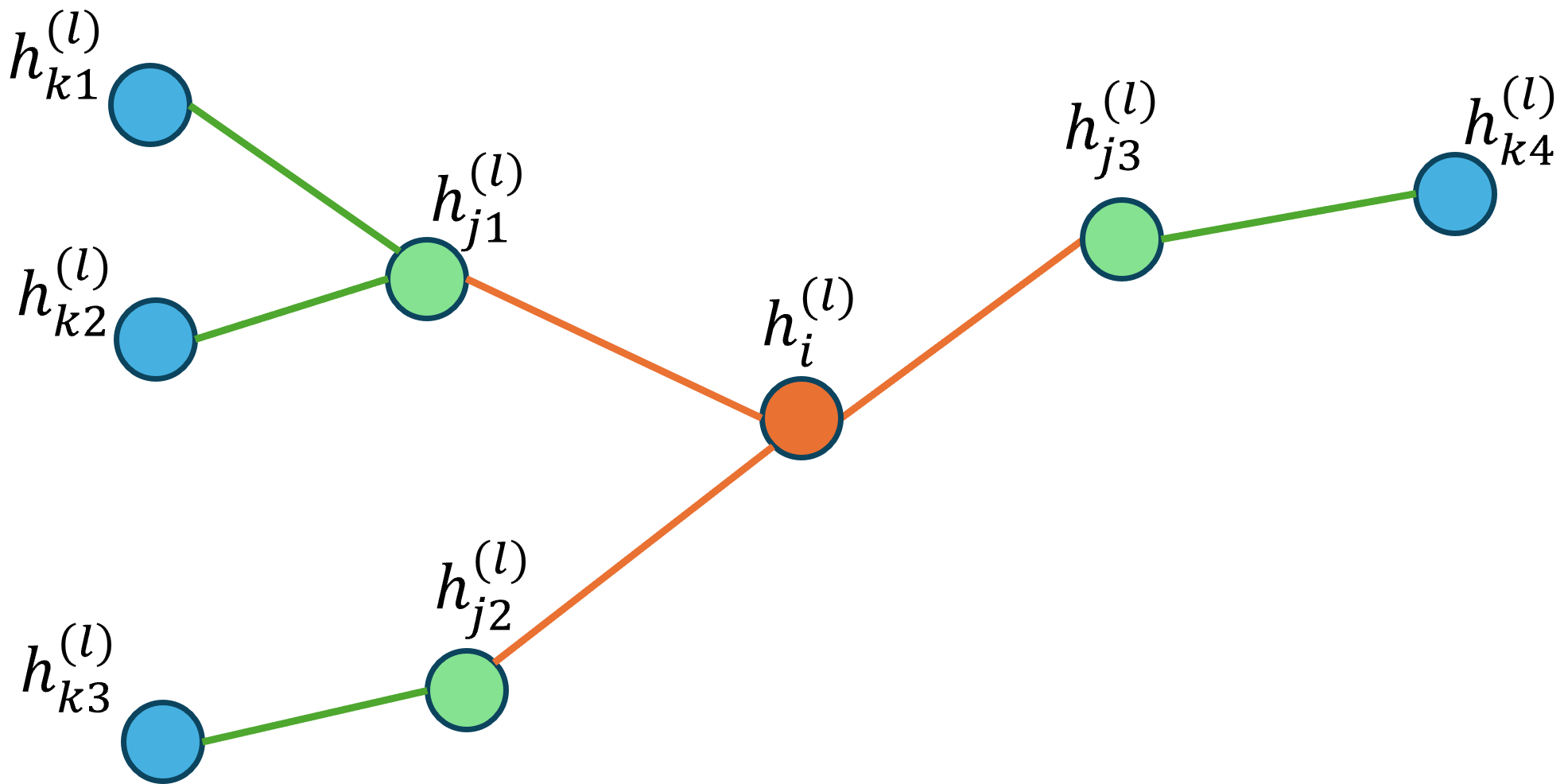}
    \caption{Illustration of two-layer message passing in a GNN. 
   }
    \label{fig:gnn_message_passing}
\end{figure}

The core operational principle of GNNs is \textit{message passing}, a framework where nodes in a graph iteratively aggregate information from their local neighbors to build increasingly sophisticated representations of themselves and their context \cite{gilmer2017neural, wu2021survey}. This process can be broken down into three steps at each layer $l$:

\begin{itemize}
    \item \textbf{Message ($m_{ij}^{(l)}$):} For every edge connecting node $j$ to node $i$, a message is computed as a function of the features of the sender node $j$, the receiver node $i$, and the edge features $e_{ij}$:
    \begin{equation}
        m_{ij}^{(l)} = \phi^{(l)} \big(h_i^{(l)}, h_j^{(l)}, e_{ij}\big)
    \end{equation}

    \item \textbf{Aggregation ($a_i^{(l)}$):} Node $i$ collects messages from its neighbors $\mathcal{N}(i)$ and combines them using a permutation-invariant aggregation function (e.g., sum, mean, max):
    \begin{equation}
        a_i^{(l)} = \bigoplus_{j \in \mathcal{N}(i)} m_{ij}^{(l)}
    \end{equation}

    \item \textbf{Update ($h_i^{(l+1)}$):} The node updates its representation by combining its previous features with the aggregated neighborhood message:
    \begin{equation}
        h_i^{(l+1)} = \psi^{(l)} \big(h_i^{(l)}, a_i^{(l)}\big)
    \end{equation}
\end{itemize}
This mechanism is illustrated in Fig.~\ref{fig:gnn_message_passing}, which depicts a 
two-layer GNN where the central node iteratively aggregates information from its neighbors and their neighbors.

This mechanism allows each node to progressively incorporate information from multi-hop neighborhoods. For example, in a 2-layer GNN, a central node aggregates features from its direct neighbors in the first layer, and from its neighbors’ neighbors in the second layer. This ability to capture higher-order structural dependencies makes GNNs particularly effective for tasks where node states are influenced by spatial and relational context \cite{kipf2017, velickovic2018gat, wu2019simplifying}.

\subsubsection*{ Problem Formulation}
Light Detection and Ranging (LiDAR) systems generate point clouds as unordered sets of points $\mathcal{P} = \{p_i\}$, where each $p_i = (x_i, y_i, z_i, r_i)$ includes 3D coordinates and reflectance intensity. These systems are prone to beam dropout, modeled by a function $\mathcal{D}(\cdot)$, which removes points from specific beams due to hardware limitations or environmental factors, resulting in a sparse subset $\mathcal{P}_{\text{sparse}} \subset \mathcal{P}$. This dropout particularly affects z-coordinates, critical for spatial perception in applications like autonomous navigation. The objective is to learn a mapping function $f_\theta$ that reconstructs the missing z-values: $\hat{z}_i = f_\theta(\mathcal{P}_{\text{sparse}})$, restoring elevation data while assuming x and y coordinates remain reliable. As illustrated in Figure \ref{fig:dropout}, this targeted reconstruction addresses a common challenge in sparse LiDAR data.

\subsubsection*{ Graph Representation}
To formulate this as a graph problem, we represent $\mathcal{P}_{\text{sparse}}$ as a graph $\mathcal{G} = (\mathcal{V}, \mathcal{E})$. Nodes $\mathcal{V}$ correspond to points $p_i$, with feature vectors $\mathbf{v}_i = [x_i, y_i, \tilde{z}_i, b_i]$, where $\tilde{z}_i$ is the observed z-coordinate (or 0 if dropped) and $b_i$ is a beam index reflecting the sensor’s vertical structure. Edges $\mathcal{E}$ are defined to capture the LiDAR’s scan pattern, connecting points within the same beam sequentially to form intra-beam relationships, and linking points across adjacent beams to model surface continuity. This approach prioritizes the sensor’s inherent geometry over pure spatial proximity, offering a flexible representation suitable for graph-based methods.

\subsubsection*{Preparation for GNN Application}
This graph structure enables the application of Graph Neural Networks (GNNs) to model local relationships among points, leveraging the beam-aware connectivity to infer missing z-values. The representation provides a foundation for advanced graph-based techniques, setting the stage for exploring convolutional and attention-based approaches to enhance reconstruction accuracy. This prepares the framework for subsequent development of tailored GNN architectures.
\subsection*{GNN Architectures for LiDAR Reconstruction}

\subsubsection*{ Graph Convolutional Networks (GCN) as Baseline}
We begin with Graph Convolutional Networks (GCNs) as a baseline approach to process the graph representation of LiDAR data. GCNs aggregate features from neighboring nodes using fixed weights based on the graph structure, enabling the model to learn spatial patterns for z-reconstruction. This method applies convolutional operations over the graph, leveraging the beam-aware edges to propagate information. While effective for initial feature extraction, GCNs’ uniform weighting limits their ability to prioritize critical local features, motivating the need for a more adaptive approach.


\subsubsection*{Graph Attention Networks: Theoretical Foundation}

Graph Attention Networks (GATs) represent a significant advancement in graph representation learning by introducing an attention mechanism that enables dynamic, content-aware neighborhood aggregation~\cite{velickovic2018}. Unlike Graph Convolutional Networks (GCNs) that employ fixed-weight aggregation based on graph structure, GATs compute adaptive attention coefficients that weight the contribution of each neighbor during feature propagation, allowing the model to focus on the most relevant contextual information for sparse LiDAR reconstruction.

The core innovation lies in the self-attention mechanism that operates on node pairs. For a given node $i$ with features $\mathbf{h}_i$ and its neighbor $j \in \mathcal{N}(i)$, the attention mechanism first applies a shared linear transformation parameterized by weight matrix $\mathbf{W} \in \mathbb{R}^{F' \times F}$:

\begin{equation}
\mathbf{h}_i' = \mathbf{W} \mathbf{h}_i, \quad \mathbf{h}_j' = \mathbf{W} \mathbf{h}_j
\end{equation}

The unnormalized attention coefficient is then computed as:

\begin{equation}
e_{ij} = \text{LeakyReLU}\left(\mathbf{a}^T[\mathbf{h}_i' \| \mathbf{h}_j']\right)
\end{equation}

where $\mathbf{a} \in \mathbb{R}^{2F'}$ is a learnable attention vector, and $\|$ denotes vector concatenation. The attention coefficients are normalized across the neighborhood using the softmax function:

\begin{equation}
\alpha_{ij} = \frac{\exp(e_{ij})}{\sum_{k \in \mathcal{N}(i)} \exp(e_{ik})}
\end{equation}

The updated node representation is computed as the attention-weighted aggregation:

\begin{equation}
\mathbf{h}_i^{\text{out}} = \sigma\left(\sum_{j \in \mathcal{N}(i)} \alpha_{ij} \mathbf{h}_j'\right)
\end{equation}

where $\sigma$ is a nonlinear activation function. This formulation enables anisotropic aggregation, where each neighbor's contribution is dynamically weighted based on feature compatibility, providing a more expressive scheme than isotropic approaches~\cite{velickovic2018,gilmer2017neural}. The standard GAT architecture follows the sequential processing pipeline illustrated in Figure~3a.

\subsection*{SuperiorGAT: Enhanced Architecture for LiDAR Reconstruction}

\begin{figure}[ht]
    \centering
    \includegraphics[width=0.75\textwidth]{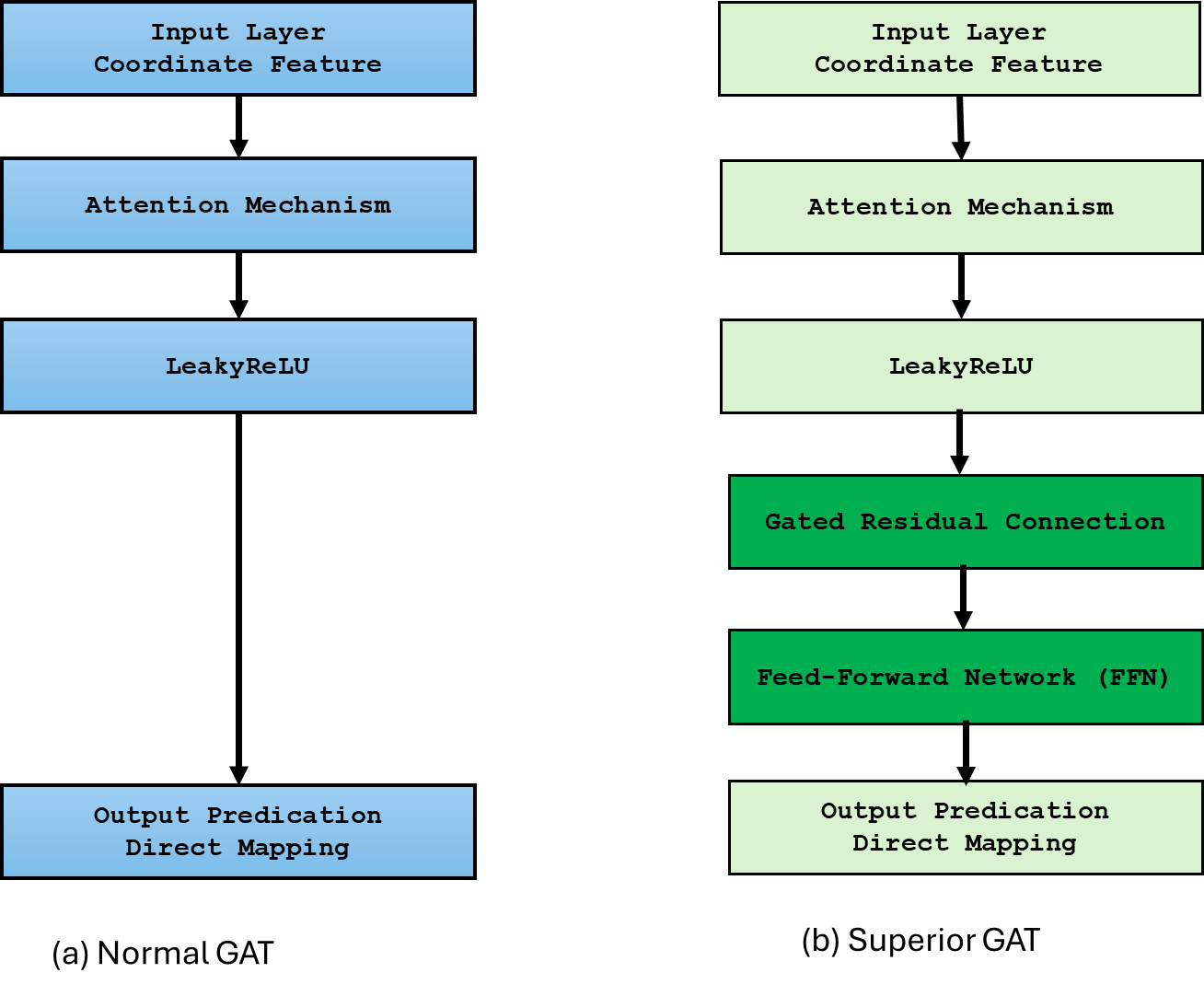}
    \caption{GAT architecture comparison. (a) Standard implementation. (b) SuperiorGAT with domain-specific enhancements for LiDAR reconstruction.}
    \label{fig:Fig3}
\end{figure}

\begin{figure}[p]
    \centering
    \includegraphics[angle=-90,origin=c,height=.7\textheight]{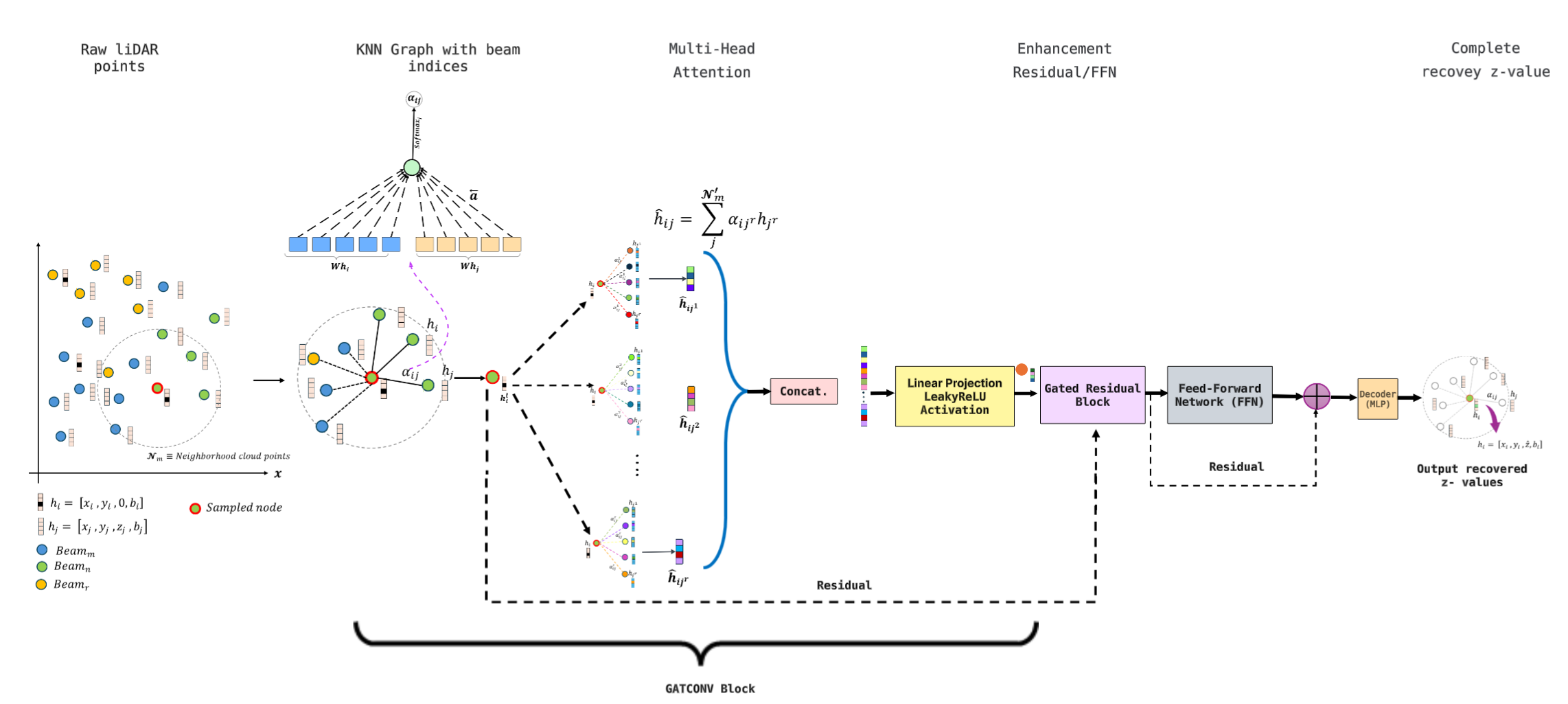}
    \caption{Beam-indexed graph structure used for LiDAR point reconstruction, illustrating intra- and inter-beam connections. Dropped points are marked to show the reconstruction target.}
    \label{fig:Fig4}
\end{figure}

While standard GATs provide adaptive neighbor weighting, they lack several components crucial for handling the distinctive sparsity patterns and irregular geometry of LiDAR point clouds. To address these limitations, we propose SuperiorGAT---a specialized architecture with targeted enhancements for reconstructing missing points in sparse 3D scans. The overall framework comparison is illustrated in Figure~\ref{fig:Fig3}, while the detailed computational pipeline is shown in Figure~\ref{fig:Fig4}.

As visualized in Figure~\ref{fig:Fig4}, SuperiorGAT processes each node through four sequential stages: (1) beam-aware feature encoding, (2) multi-head attention aggregation, (3) gated residual fusion, and (4) feed-forward refinement with task-specific decoding.

\textbf{Beam-Aware Feature Encoding.}  
SuperiorGAT begins by enriching each node with structural information from the LiDAR sensor configuration. Rather than relying solely on spatial coordinates, each point $i$ is represented with scan-pattern context:
\begin{equation}
\mathbf{h}_i^{(0)} = [x_i,\, y_i,\, \tilde{z}_i,\, b_i],
\label{eq:beam_encoding}
\end{equation}
where $b_i$ denotes the beam index. This encoding provides critical geometric context about the LiDAR scanning mechanism and significantly enhances the model's capacity to reason about missing depths resulting from beam dropout.

\textbf{Multi-Head Attention with Adaptive Aggregation.}  
The encoded features undergo multi-head graph attention to capture diverse geometric relationships. For each attention head $k$, the node representation is updated as:
\begin{equation}
\mathbf{h}_i^{\text{attn},k} = \sigma\left( \sum_{j \in \mathcal{N}(i)} \alpha_{ij}^k \mathbf{W}^k \mathbf{h}_j^{(0)} \right),
\end{equation}
where $\alpha_{ij}^k$ are the attention coefficients from head $k$, $\mathbf{W}^k$ is the head-specific weight matrix, and $\sigma$ denotes the activation function. The outputs from all $K$ heads are concatenated:
\begin{equation}
\mathbf{h}_i^{\text{attn}} = \bigg\|_{k=1}^{K} \mathbf{h}_i^{\text{attn},k}.
\end{equation}

where $\mathbf{h}_{i}^{\text{attn},k} \in \mathbb{R}^{F'}$ is the output of head $k$,
so that $\mathbf{h}_i^{\text{attn}} \in \mathbb{R}^{K F'}$ collects all heads into a
single feature vector.

\textbf{Stabilized Attention Through Residual Gating.}  
Following attention aggregation, SuperiorGAT employs a gated residual mechanism that dynamically balances the attention-refined features with the original normalized input:
\begin{equation}
\mathbf{h}_i^{\text{gated}} = \text{LayerNorm}\!\left( \gamma \cdot \mathbf{h}_i^{\text{attn}} + (1 - \gamma) \cdot \mathbf{h}_i^{\text{norm}} \right),
\label{eq:gated_residual}
\end{equation}
where $\gamma \in [0,1]$ is a learnable gate parameter and $\mathbf{h}_i^{\text{norm}}$ represents the normalized input features. This adaptive gating mechanism provides crucial stability in regions where neighborhood information may be unreliable due to sparse sampling or beam dropout.

\textbf{Feed-Forward Refinement.}  
The gated representations then pass through a compact feed-forward network (FFN) for non-linear feature enhancement. A residual connection preserves structural information throughout this refinement:
\begin{equation}
\mathbf{h}_i^{\text{final}} = \text{LayerNorm}\!\left( \text{FFN}(\mathbf{h}_i^{\text{gated}}) + \mathbf{h}_i^{\text{gated}} \right),
\label{eq:ffn_refinement}
\end{equation}
where $\text{FFN}(\mathbf{x}) = \mathbf{W}_2 \cdot \text{LeakyReLU}(\mathbf{W}_1 \mathbf{x} + \mathbf{b}_1) + \mathbf{b}_2$. This expansion-contraction design enhances representational capacity without significantly increasing computational complexity.

\textbf{Task-Specific Output Decoding.}  
The final refined feature vector $\mathbf{h}_i^{\text{final}}$ is processed by a lightweight multi-layer perceptron optimized specifically for elevation recovery:
\begin{equation}
\hat{z}_i = \text{MLP}_{\text{decoder}}(\mathbf{h}_i^{\text{final}}).
\end{equation}
This regression-focused output head predicts the reconstructed depth value $\hat{z}_i$ for each point, completing the restoration of missing elevation data.

The integrated combination of beam-index encoding, multi-head attention with adaptive aggregation, gated residual stabilization, FFN refinement, and specialized decoding enables SuperiorGAT to effectively address the distinctive challenges of LiDAR point cloud reconstruction under beam dropout conditions. As illustrated in Figure ~\ref{fig:Fig4}, SuperiorGAT uses two distinct residual pathways: one before the gated fusion and another inside the FFN block, ensuring stable propagation of geometric information.

\section*{Experimental Setup}

All experiments were conducted using two public autonomous driving LiDAR datasets: the KITTI raw dataset \cite{Geiger2013IJRR} and the nuScenes dataset \cite{nuScene}. These datasets were selected to evaluate the proposed method under different sensor configurations and scene characteristics.

For the KITTI dataset, experiments focused on a residential driving sequence recorded with a Velodyne HDL-64E sensor operating at 10~Hz. The selected sequence contains 200 frames and represents a typical urban residential scenario. Each LiDAR scan provides three-dimensional point coordinates, from which approximately 50{,}000 points per frame were retained using stratified sampling to preserve the vertical beam structure.

In addition to the KITTI dataset, we evaluated the proposed framework on the nuScenes dataset \cite{nuScene} to assess robustness across LiDAR sensors with differing vertical resolutions. nuScenes is a large-scale autonomous driving dataset collected in urban environments using a Velodyne HDL-32E LiDAR sensor operating at 20 Hz, which provides 32 vertical scanning beams, half the vertical resolution of the KITTI sensor.

For nuScenes experiments, LiDAR frames were processed using the same pipeline applied to KITTI, with beam-aware graph construction and per-frame independent reconstruction. Due to the reduced number of vertical beams, structured beam dropout was applied proportionally by removing fixed subsets of beams to simulate degradation rates of 12.5\%, 25\%, and 37.5\%. This setting reflects realistic failure or sparsification scenarios commonly encountered in lower-cost LiDAR systems.

All models were evaluated using identical graph parameters and architectural configurations to those used in the KITTI experiments, including the same neighborhood size \(k\). Performance metrics were computed exclusively on the dropped points using RMSE in the vertical direction and geometric consistency measures. This cross-dataset evaluation allows direct comparison of reconstruction behavior under varying sensor resolutions while maintaining a consistent experimental protocol. Models are trained and evaluated independently on each dataset to ensure fair comparison under dataset-specific sensor configurations, without cross-dataset fine-tuning.

\section*{Results and Discussion}

\subsection*{Neighborhood Sensitivity Analysis}

\begin{figure}[H]
    \centering
    \includegraphics[width=\linewidth]{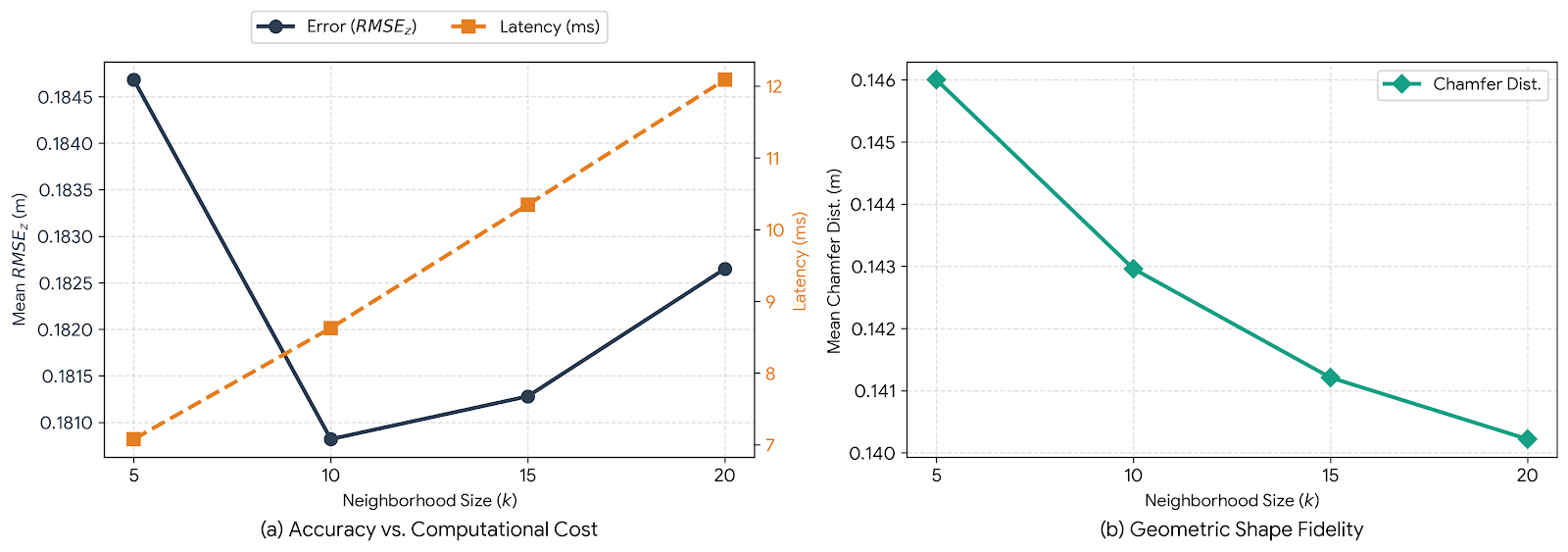}
    \caption{Sensitivity analysis of the neighborhood size $k$ showing (a) the trade-off between vertical reconstruction error ($RMSE_z$) and inference latency, and (b) the corresponding geometric fidelity measured by Chamfer distance.}
    \label{fig:Fig05new}
\end{figure}

We first analyze the sensitivity of the proposed SuperiorGAT model to the neighborhood size parameter $k$, which determines the number of neighboring points used to construct the local graph for attention-based aggregation. This parameter directly controls the balance between local geometric detail and broader spatial context. The analysis is conducted on the KITTI Residential dataset, and the results are illustrated in Fig.~\ref{fig:Fig05new}.

As shown in Fig.~\ref{fig:Fig05new}(a), increasing the neighborhood size from $k=5$ to $k=10$ leads to a significant reduction in vertical reconstruction error, achieving the lowest $RMSE_z$ of 0.1808\,m at $k=10$. Beyond this point, further increasing $k$ results in a gradual increase in $RMSE_z$, indicating that excessive neighborhood aggregation introduces over-smoothing effects that degrade local vertical precision.

In contrast, Fig.~\ref{fig:Fig05new}(b) shows that the Chamfer distance continues to decrease as $k$ increases, reflecting improved global geometric alignment when larger spatial contexts are considered. This divergence between vertical accuracy and global shape fidelity highlights an inherent trade-off: smaller neighborhoods preserve fine-grained elevation details, while larger neighborhoods favor smoother but less locally precise reconstructions.

From a computational perspective, inference latency increases approximately linearly with $k$, rising from 7.08\,ms at $k=5$ to 12.09\,ms at $k=20$. Considering reconstruction accuracy, geometric fidelity, and runtime jointly, $k=10$ represents a clear efficiency–accuracy balance point. Consequently, $k=10$ is selected as the default configuration and is used in all subsequent experiments.

\subsection*{Cross-Environment Evaluation on KITTI}

After fixing the neighborhood size to $k=10$, we evaluate the generalization capability of SuperiorGAT across multiple real-world operating conditions using five distinct KITTI sequences: \textbf{Residential, City, Person, Campus, and Road}. These environments exhibit substantial variability in scene structure, object density, and terrain complexity, providing a rigorous test of cross-domain robustness.

Quantitative results are summarized in Table~1. SuperiorGAT consistently achieves strong reconstruction performance across all environments, maintaining an $RMSE_{xyz}$ below 0.11\,m in every sequence while simultaneously producing the lowest or near-lowest Chamfer distances among learning-based methods. This indicates that the proposed model effectively preserves both vertical accuracy and global geometric structure across diverse scene types.

In contrast, Nearest Neighbor interpolation occasionally yields competitive $RMSE_z$ values in specific environments; however, this apparent advantage does not translate to improved three-dimensional fidelity. The corresponding Chamfer distances are substantially higher, reflecting poor global alignment and the presence of structural artifacts. This behavior highlights the limitation of purely local interpolation schemes, which fail to model spatial context beyond immediate neighbors.

Compared to the standard GAT baseline, SuperiorGAT demonstrates improved reconstruction accuracy while reducing inference latency from approximately 13\,ms to 8.6\,ms per frame. Importantly, this latency remains stable across all evaluated environments, indicating that the proposed architecture does not rely on environment-specific tuning. Overall, these results confirm that SuperiorGAT generalizes reliably across heterogeneous KITTI scenes while maintaining inference latency compatible with real-time autonomous perception pipelines. This efficiency gain is achieved by avoiding stacked attention layers and instead employing a single beam-aware graph attention layer augmented with gated residual fusion and a lightweight feed-forward refinement block, which reduces computational overhead without sacrificing reconstruction fidelity.

\begin{table*}[!t]
\centering
\caption{Performance Comparison Across Different KITTI Environments (Mean $\pm$ SD)}
\label{tab:generalization_results}
\resizebox{\textwidth}{!}{%
\begin{tabular}{l l c c c c}
\hline
Dataset & Method & $RMSE_z$ (m) & $RMSE_{xyz}$ (m) & Chamfer Distance (m) & Time (ms) \\ \hline
& Linear Interp & 0.472 $\pm$ 0.102 & 0.272 $\pm$ 0.059 & 0.202 $\pm$ 0.015 & 0.51 $\pm$ 0.02 \\
& Nearest Neighbor & 0.174 $\pm$ 0.065 & 0.330 $\pm$ 0.031 & 0.546 $\pm$ 0.019 & 0.05 $\pm$ 0.00 \\
\textbf{Residential} & Enhanced PointNet & 0.499 $\pm$ 0.132 & 0.288 $\pm$ 0.076 & 0.603 $\pm$ 0.178 & 3.57 $\pm$ 0.06 \\
& Simple GCN & 0.258 $\pm$ 0.056 & 0.149 $\pm$ 0.032 & 0.244 $\pm$ 0.016 & 7.18 $\pm$ 0.05 \\
& GAT Baseline & 0.193 $\pm$ 0.054 & 0.112 $\pm$ 0.031 & 0.170 $\pm$ 0.008 & 13.08 $\pm$ 0.28 \\
& \textbf{SuperiorGAT (Ours)} & 0.181 $\pm$ 0.057 & \textbf{0.104 $\pm$ 0.033} & \textbf{0.143 $\pm$ 0.008} & 8.63 $\pm$ 0.06 \\
\hline
& Linear Interp & 0.437 $\pm$ 0.048 & 0.252 $\pm$ 0.028 & 0.213 $\pm$ 0.025 & 0.56 $\pm$ 0.03 \\
& Nearest Neighbor & 0.131 $\pm$ 0.026 & 0.224 $\pm$ 0.048 & 0.373 $\pm$ 0.078 & 0.05 $\pm$ 0.00 \\
\textbf{City} & Enhanced PointNet & 0.461 $\pm$ 0.109 & 0.266 $\pm$ 0.063 & 0.511 $\pm$ 0.133 & 3.64 $\pm$ 0.08 \\
& Simple GCN & 0.307 $\pm$ 0.024 & 0.177 $\pm$ 0.014 & 0.327 $\pm$ 0.033 & 7.26 $\pm$ 0.10 \\
& GAT Baseline & 0.229 $\pm$ 0.034 & 0.132 $\pm$ 0.020 & 0.244 $\pm$ 0.037 & 13.23 $\pm$ 0.09 \\
& \textbf{SuperiorGAT (Ours)} & 0.152 $\pm$ 0.029 & \textbf{0.088 $\pm$ 0.017} & \textbf{0.147 $\pm$ 0.035} & 8.70 $\pm$ 0.08 \\
\hline
& Linear Interp & 0.470 $\pm$ 0.009 & 0.271 $\pm$ 0.005 & 0.208 $\pm$ 0.004 & 0.53 $\pm$ 0.02 \\
& Nearest Neighbor & 0.110 $\pm$ 0.008 & 0.229 $\pm$ 0.014 & 0.373 $\pm$ 0.006 & 0.05 $\pm$ 0.00 \\
\textbf{Person} & Enhanced PointNet & 0.350 $\pm$ 0.058 & 0.202 $\pm$ 0.033 & 0.417 $\pm$ 0.073 & 3.56 $\pm$ 0.03 \\
& Simple GCN & 0.252 $\pm$ 0.011 & 0.145 $\pm$ 0.007 & 0.262 $\pm$ 0.011 & 7.14 $\pm$ 0.04 \\
& GAT Baseline & 0.211 $\pm$ 0.010 & 0.122 $\pm$ 0.006 & 0.217 $\pm$ 0.009 & 13.10 $\pm$ 0.03 \\
& \textbf{SuperiorGAT (Ours)} & 0.147 $\pm$ 0.011 & \textbf{0.085 $\pm$ 0.006} & \textbf{0.139 $\pm$ 0.013} & 8.58 $\pm$ 0.03 \\
\hline
& Linear Interp & 0.412 $\pm$ 0.041 & 0.238 $\pm$ 0.024 & 0.217 $\pm$ 0.015 & 0.52 $\pm$ 0.02 \\
& Nearest Neighbor & 0.146 $\pm$ 0.011 & 0.265 $\pm$ 0.036 & 0.456 $\pm$ 0.009 & 0.05 $\pm$ 0.00 \\
\textbf{Campus} & Enhanced PointNet & 0.433 $\pm$ 0.103 & 0.250 $\pm$ 0.059 & 0.497 $\pm$ 0.128 & 3.67 $\pm$ 0.05 \\
& Simple GCN & 0.254 $\pm$ 0.023 & 0.147 $\pm$ 0.013 & 0.252 $\pm$ 0.018 & 7.29 $\pm$ 0.07 \\
& GAT Baseline & 0.211 $\pm$ 0.017 & 0.122 $\pm$ 0.010 & 0.205 $\pm$ 0.016 & 13.25 $\pm$ 0.10 \\
& \textbf{SuperiorGAT (Ours)} & 0.166 $\pm$ 0.019 & \textbf{0.096 $\pm$ 0.011} & \textbf{0.149 $\pm$ 0.011} & 8.74 $\pm$ 0.05 \\
\hline
& Linear Interp & 0.470 $\pm$ 0.009 & 0.271 $\pm$ 0.005 & 0.208 $\pm$ 0.004 & 0.52 $\pm$ 0.01 \\
& Nearest Neighbor & 0.110 $\pm$ 0.008 & 0.229 $\pm$ 0.014 & 0.373 $\pm$ 0.006 & 0.05 $\pm$ 0.00 \\
\textbf{Road} & Enhanced PointNet & 0.350 $\pm$ 0.058 & 0.202 $\pm$ 0.033 & 0.417 $\pm$ 0.073 & 3.57 $\pm$ 0.03 \\
& Simple GCN & 0.252 $\pm$ 0.011 & 0.145 $\pm$ 0.007 & 0.262 $\pm$ 0.011 & 7.14 $\pm$ 0.04 \\
& GAT Baseline & 0.211 $\pm$ 0.010 & 0.122 $\pm$ 0.006 & 0.217 $\pm$ 0.009 & 13.11 $\pm$ 0.04 \\
& \textbf{SuperiorGAT (Ours)} & 0.147 $\pm$ 0.011 & \textbf{0.085 $\pm$ 0.006} & \textbf{0.139 $\pm$ 0.013} & 8.59 $\pm$ 0.03 \\
\hline
\end{tabular}%
}
\end{table*}

\subsection*{Robustness to Structured Beam Dropout}

To assess robustness under realistic sensor degradation, we evaluate SuperiorGAT under structured beam dropout on both the KITTI and nuScenes datasets. Unlike random point removal, structured beam dropout eliminates contiguous vertical scan lines, emulating reduced-resolution LiDAR sensors, partial sensor failure, or adverse environmental conditions. Dropout rates of 12.5\%, 25\%, and 37.5\% are considered to progressively increase reconstruction difficulty.

\begin{figure}[t]
    \centering
    \begin{subfigure}[b]{0.48\linewidth}
        \centering
        \includegraphics[width=\linewidth]{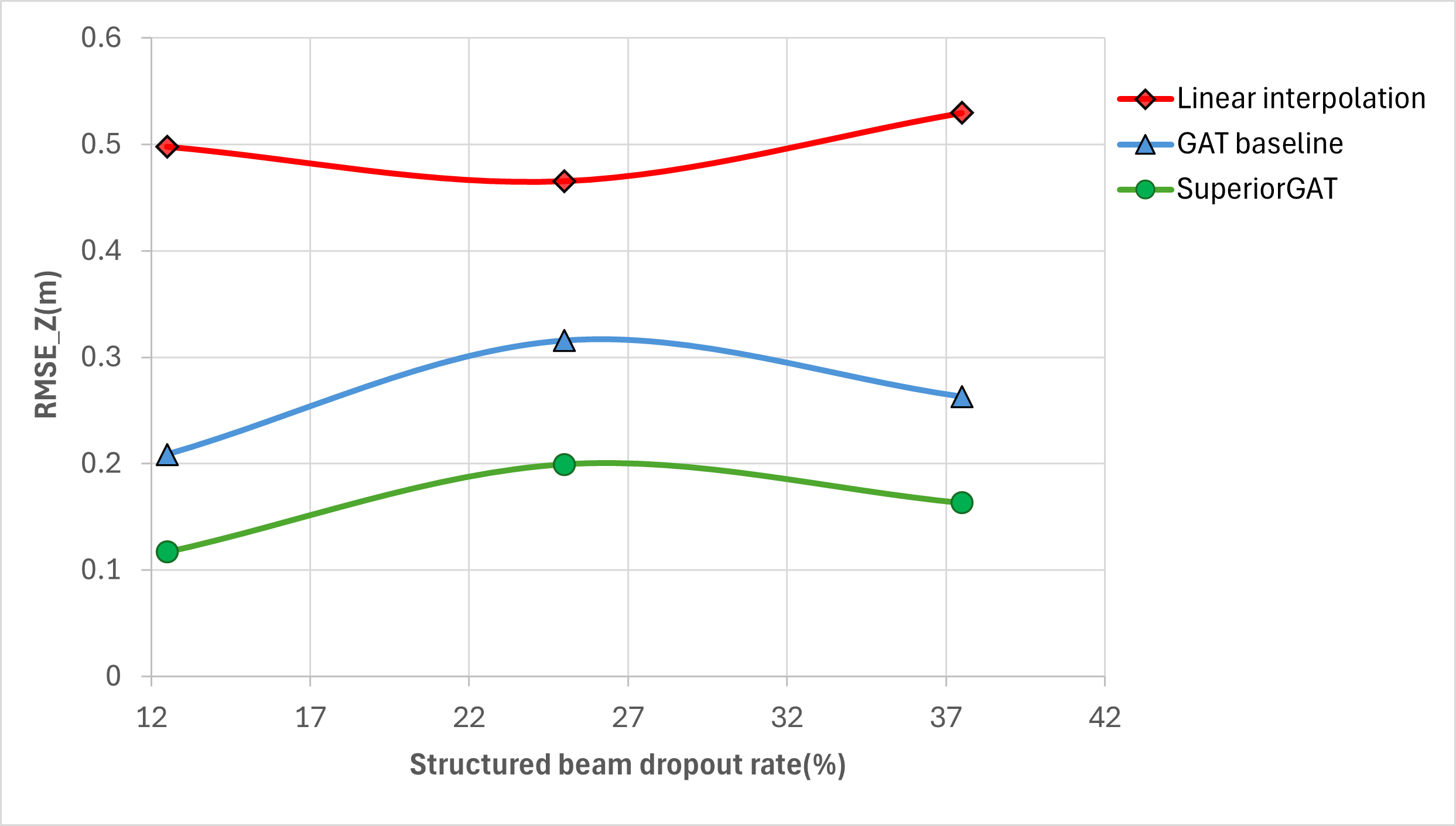}
        \caption{KITTI dataset}
        \label{fig:dropout_kitti}
    \end{subfigure}
    \hfill
    \begin{subfigure}[b]{0.48\linewidth}
        \centering
        \includegraphics[width=\linewidth]{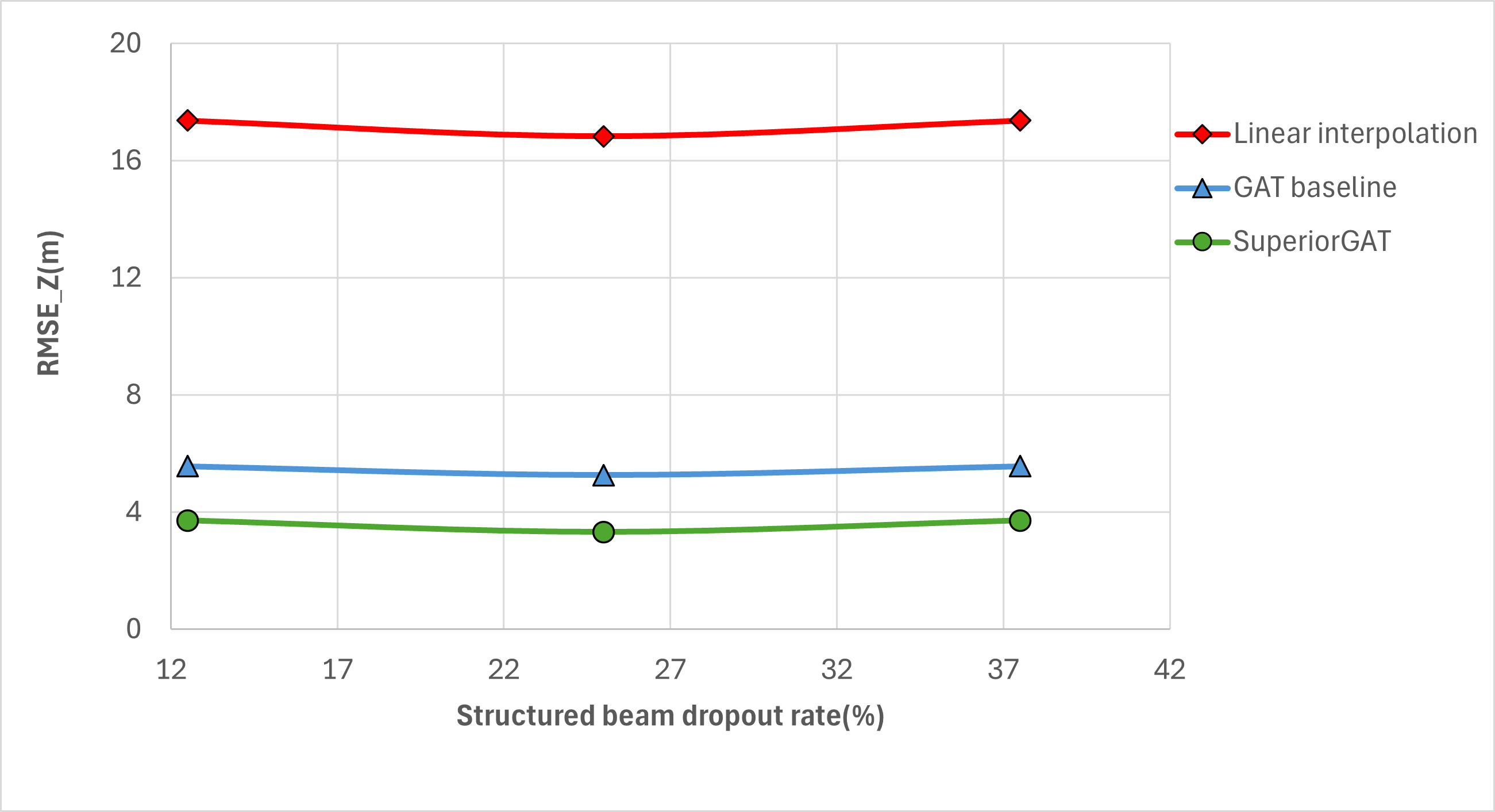}
        \caption{nuScenes dataset}
        \label{fig:dropout_nuscenes}
    \end{subfigure}
    \caption{Vertical reconstruction error ($RMSE_z$) as a function of structured beam dropout rate for (a) KITTI and (b) nuScenes. SuperiorGAT exhibits consistently lower error and more graceful degradation compared to baseline methods across increasing dropout levels.}
    \label{fig:dropout_robustness}
\end{figure}

Figs.~\ref{fig:dropout_robustness}a and~\ref{fig:dropout_robustness}b
 illustrate the effect of increasing beam dropout on vertical reconstruction accuracy for KITTI and nuScenes, respectively. Across both datasets, SuperiorGAT consistently achieves the lowest $RMSE_z$ at all dropout levels and exhibits a significantly slower degradation rate compared to baseline methods. This behavior indicates that the proposed attention-based architecture effectively leverages remaining spatial context to recover missing vertical structure even under severe information loss.

Quantitative results are further detailed in Tables~2 and~3. On KITTI, SuperiorGAT maintains stable performance as dropout increases, with only moderate growth in $RMSE_z$ and Chamfer distance, while Linear Interpolation shows large and largely invariant errors regardless of dropout severity. The GAT baseline degrades more rapidly, highlighting its limited ability to adapt to structured beam removal.

A similar trend is observed on nuScenes, which operates with a lower vertical resolution 32-beam LiDAR sensor. Despite the inherently more challenging reconstruction setting, SuperiorGAT preserves a clear performance margin over both interpolation-based and graph-based baselines across all dropout rates. Notably, inference time remains nearly constant for all dropout configurations on both datasets, confirming that robustness is achieved through learned spatial reasoning rather than increased computational cost.

Overall, these results demonstrate that SuperiorGAT degrades gracefully under structured beam dropout and generalizes robustly across datasets with different LiDAR configurations, reinforcing its suitability for deployment in real-world autonomous perception scenarios.

\begin{table}[ht]
\centering
\caption{Robustness under structured beam dropout on KITTI (mean over frames, averaged over three runs).}
\label{tab:kitti_dropout}
\begin{tabular}{c|l|cccc}
\hline
Dropout &
Method &
RMSE$_Z$ (m) $\downarrow$ &
Chamfer (m) $\downarrow$ &
Normal Consistency $\uparrow$ &
Inference Time (s) $\downarrow$ \\
\hline
\multirow{3}{*}{12.5\%}
& Linear Interp & 0.50 & 0.20 & 0.81 & 0.03 \\
& GAT Baseline & 0.21 & 0.17 & 0.89 & 0.04 \\
& \textbf{SuperiorGAT (Ours)} & \textbf{0.12} & \textbf{0.15} & \textbf{0.88} & \textbf{0.02} \\
\hline
\multirow{3}{*}{25\%}
& Linear Interp & 0.47 & 0.20 & 0.80 & 0.03 \\
& GAT Baseline & 0.31 & 0.24 & 0.85 & 0.04 \\
& \textbf{SuperiorGAT (Ours)} & \textbf{0.20} & \textbf{0.15} & \textbf{0.85} & \textbf{0.02} \\
\hline
\multirow{3}{*}{37.5\%}
& Linear Interp & 0.53 & 0.21 & 0.78 & 0.03 \\
& GAT Baseline & 0.26 & 0.26 & 0.83 & 0.04 \\
& \textbf{SuperiorGAT (Ours)} & \textbf{0.16} & \textbf{0.16} & \textbf{0.84} & \textbf{0.02} \\
\hline
\end{tabular}
\end{table}

\begin{table}[ht]
\centering
\caption{Robustness under structured beam dropout on nuScenes (mean over frames, averaged over three runs).}
\label{tab:nuscenes_dropout}
\begin{tabular}{c|l|cccc}
\hline
Dropout &
Method &
RMSE$_Z$ (m) $\downarrow$ &
Chamfer (m) $\downarrow$ &
Normal Consistency $\uparrow$ &
Inference Time (s) $\downarrow$ \\
\hline
\multirow{3}{*}{12.5\%}
& Linear Interp & 16.83 & 9.15 & 0.49 & 0.03 \\
& GAT Baseline & 5.27 & 2.63 & 0.68 & 0.04 \\
& \textbf{SuperiorGAT (Ours)} & \textbf{3.32} & \textbf{1.37} & \textbf{0.75} & \textbf{0.02} \\
\hline
\multirow{3}{*}{25\%}
& Linear Interp & 16.84 & 9.15 & 0.49 & 0.03 \\
& GAT Baseline & 5.30 & 2.65 & 0.68 & 0.04 \\
& \textbf{SuperiorGAT (Ours)} & \textbf{3.35} & \textbf{1.40} & \textbf{0.74} & \textbf{0.02} \\
\hline
\multirow{3}{*}{37.5\%}
& Linear Interp & 17.25 & 9.32 & 0.61 & 0.03 \\
& GAT Baseline & 5.60 & 2.90 & 0.64 & 0.04 \\
& \textbf{SuperiorGAT (Ours)} & \textbf{3.75} & \textbf{1.55} & \textbf{0.72} & \textbf{0.02} \\
\hline
\end{tabular}
\end{table}

\subsection*{Qualitative Reconstruction Analysis}

\begin{figure}[!t]
    \centering
    \includegraphics[width=1.0\textwidth]{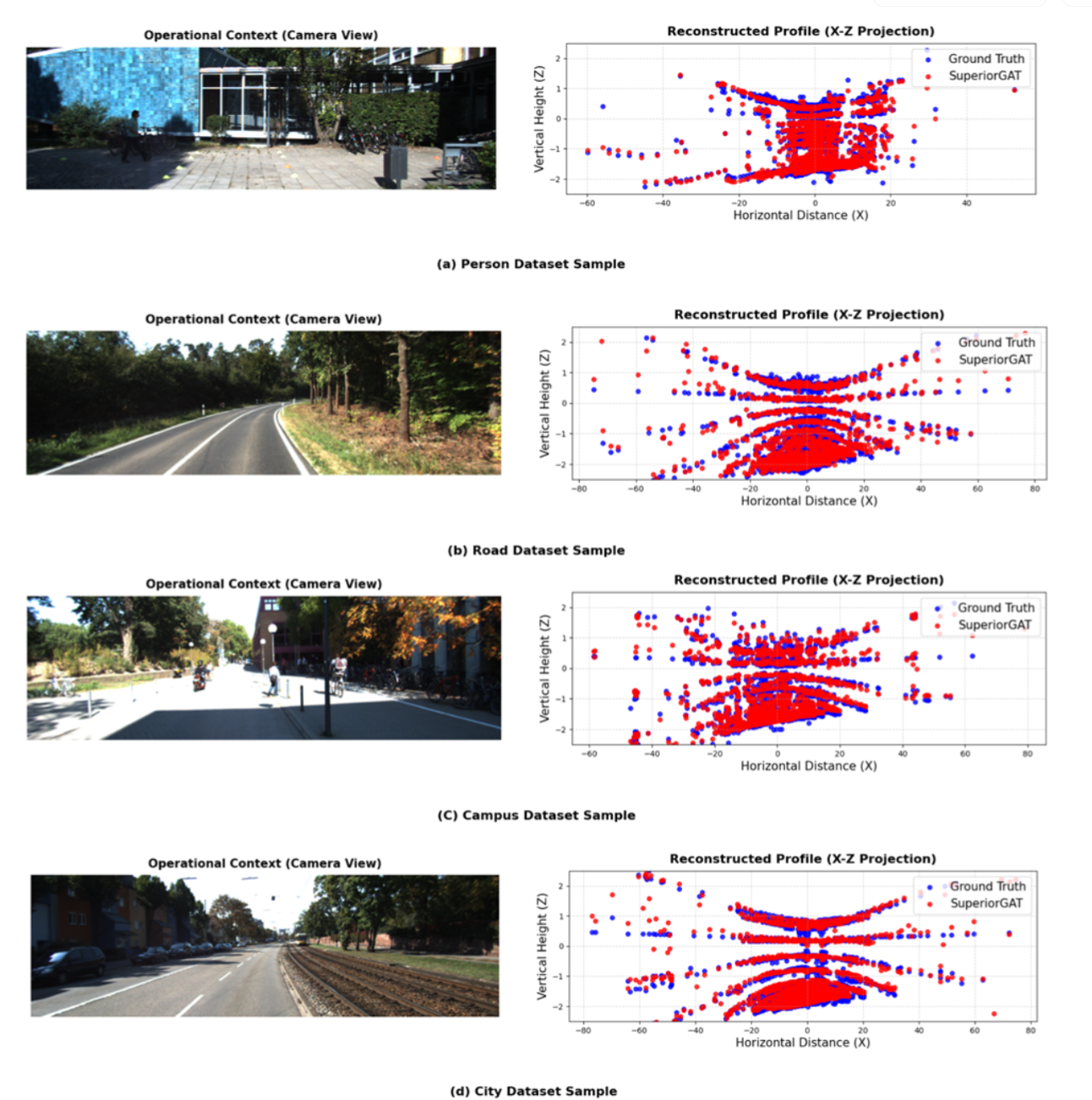}
    \caption{Qualitative reconstruction results across four datasets: (a) Person, (b) Road, (c) Campus, and (d) City. Red points (SuperiorGAT) demonstrate high vertical alignment with blue points (Ground Truth) in the X-Z projection.}
    \label{fig:Fig07}
\end{figure}

Figure~\ref{fig:Fig07} presents a qualitative comparison between the original ground-truth LiDAR data and the reconstructed vertical profiles generated by SuperiorGAT across the \textbf{Person, Road, Campus, and City} datasets. For each scene, a camera-view reference is shown on the left to provide operational context, while the corresponding 2D X--Z projection is displayed on the right, illustrating the reconstructed vertical structure as a function of horizontal distance.

For visual clarity, the reconstructed point clouds are displayed using a sampling stride of 15. Despite this downsampling, SuperiorGAT preserves high geometric fidelity, accurately recovering vertical structural height ($Z$) while maintaining smooth and coherent profiles along the horizontal axis ($X$). Across all evaluated environments, the reconstructed profiles closely follow the ground-truth geometry, capturing elevation changes, slopes, and terrain undulations without introducing spurious discontinuities.

In contrast to interpolation-based methods, which typically produce staircase artifacts and fragmented vertical profiles, SuperiorGAT maintains consistent vertical alignment and continuity. These visual results further confirm the model’s ability to integrate local and global spatial context, complementing the quantitative improvements reported in Tables~1--3 and the robustness trends observed in Figs.~\ref{fig:dropout_robustness}a and~\ref{fig:dropout_robustness}b.

Overall, the qualitative comparisons reinforce the numerical findings and demonstrate that SuperiorGAT achieves robust and geometrically consistent vertical reconstruction across diverse environments and structured beam dropout conditions.

\clearpage

\section*{Conclusion}

This paper presented \textbf{SuperiorGAT}, a graph attention--based framework for reconstructing missing elevation information in sparse LiDAR point clouds under structured beam dropout. By representing LiDAR scans as beam-aware graphs and augmenting standard graph attention networks with gated residual fusion and a lightweight feed-forward refinement module, the proposed approach effectively recovers lost vertical geometry while maintaining computational efficiency without increasing network depth.

Extensive experiments across multiple KITTI environments, including \textbf{Person}, \textbf{Road}, \textbf{Campus}, and \textbf{City}, demonstrate that SuperiorGAT consistently achieves lower reconstruction error and improved geometric consistency compared to traditional interpolation techniques, PointNet-based models, and deeper GAT baselines. Robustness evaluations under increasing structured beam dropout further show that SuperiorGAT degrades gracefully and preserves a clear performance margin across both KITTI and nuScenes datasets, highlighting its ability to generalize across different LiDAR configurations and sensing resolutions.

Qualitative analyses based on X--Z projection profiles corroborate the quantitative findings, confirming that SuperiorGAT preserves structural continuity and vertical alignment across diverse scenes with minimal distortion. As the proposed framework operates solely on geometric information and does not rely on temporal aggregation or RGB data, it offers a practical and scalable solution for LiDAR reconstruction in autonomous perception pipelines. Future work will explore extensions to dynamic dropout patterns and temporal integration to further enhance reconstruction stability in highly dynamic environments. Future work will investigate component-level ablation and cross-dataset generalization to further characterize the model’s robustness under diverse sensor configurations and dynamic dropout patterns.

\bibliography{sample}

\end{document}